\documentclass[10pt,journal,twocolumn]{IEEEtran}

\usepackage{graphics}
\usepackage{graphicx}
\usepackage{amsmath,amssymb} 
\usepackage{color}
\usepackage{bm}
\usepackage{caption}
\usepackage{multirow}
\usepackage{tabulary}
\usepackage{tabularx}
\usepackage{subcaption}
\usepackage{epsfig}
\usepackage{times}
\usepackage{epsfig}
\usepackage[Symbol]{upgreek}
\usepackage{subcaption}
\usepackage{booktabs}
\usepackage[ruled]{algorithm2e}

\usepackage[mathscr]{euscript}

\usepackage[pagebackref=true,breaklinks=true,letterpaper=true,colorlinks,bookmarks=false]{hyperref}

\ifCLASSINFOpdf
\else
\fi
\begin{document}
\title{Adaptive Algorithm and Platform Selection for Visual Detection and Tracking}

%\author{Shu Zhang,~\IEEEmembership{Member,~IEEE,}
%        John~Doe,~\IEEEmembership{Fellow,~OSA,}
%        and~Jane~Doe,~\IEEEmembership{Life~Fellow,~IEEE}% <-this % stops a space
\author{Shu Zhang,
       Qi Zhu,
        and~Amit K. Roy-Chowdhury % <-this
%\thanks{Shu Zhang is with the Department
%of Electrical Engineering, University of California, Riverside,
%CA, 92521 USA e-mail: (szhang@ee.ucr.edu).}% <-this % stops a space
%\thanks{Manuscript received April 19, 2005; revised January 11, 2007.}

\thanks{Shu Zhang, Qi Zhu, and Amit K. Roy-Chowdhury are with the Department of Electrical and Computer Engineering, University of California, Riverside, CA, USA, 92521.}

}



\maketitle

\begin{abstract}

Computer vision algorithms are known to be extremely sensitive to the environmental conditions in which the data is captured, e.g., lighting conditions and target density. Tuning of parameters or choosing a completely new algorithm is often needed to achieve a certain performance level, especially when there is a limitation of the computation source. In this paper, we focus on this problem and propose a framework to adaptively select the ``best'' algorithm-parameter combination (referred to as the best algorithm for simplicity) and the computation platform under performance and cost constraints at design time, and adapt the algorithms at runtime based on real-time inputs. This necessitates developing a mechanism to switch between different algorithms as the nature of the input video changes. Our proposed algorithm calculates a similarity function between a test video scenario and each unique training scenario, where the similarity calculation is based on learning a manifold of image features that is shared by both the training and test datasets. Similarity between training and test dataset indicates the same algorithm can be applied to both of them and achieve similar performance. We design a cost function with this similarity measure to find the most similar training scenario to the test data. The ``best'' algorithm under a given platform is obtained by selecting the algorithm with a specific parameter combination that performs the best on the corresponding training data. The proposed framework can be used first offline to choose the platform based on performance and cost constraints, and then online whereby the ``best'' algorithm is selected for each new incoming video segment for a given platform.
In the experiments, we apply our algorithm to the problems of pedestrian detection and tracking. We show how to adaptively select platforms and algorithm-parameter combinations. Our results provide optimal performance on 3 publicly available datasets.

\end{abstract}


% \fi
\IEEEpeerreviewmaketitle

\section{Introduction}

\IEEEPARstart{N}{umerous} algorithms have been developed for different computer vision applications like object detection, object recognition, tracking, etc. Also, many public datasets have been released to help researchers fairly evaluate their algorithms.  For instance, the datasets of CAVIAR \cite{CAVIARDataset}, ETHMS \cite{Ess08}, and TUD-Brussels \cite{Wojek09} have been commonly used in the area of tracking. In most cases, each algorithm is able to achieve very good performance on some datasets, while failing to beat other algorithms on some other datasets. Besides, it is interesting to see that some algorithms perform well on parts of a dataset, but cannot achieve good results on some other parts. This is because every algorithm is sensitive to the environmental conditions in each dataset or parts thereof.
Moreover, although some state-of-the-art algorithms can achieve better results than other algorithms in simulation, the high computation complexity might significantly reduce their performance in the real-world scenarios or requires computation platforms that are out of monetary or energy budget. In which case, choosing other algorithms with a different platform may be more beneficial.
All these observations raise an important question: can we automatically select the best computation platform and the best algorithm-parameter combination for an application domain?

\begin{figure*}
\begin{center}

\includegraphics[width=1\linewidth]{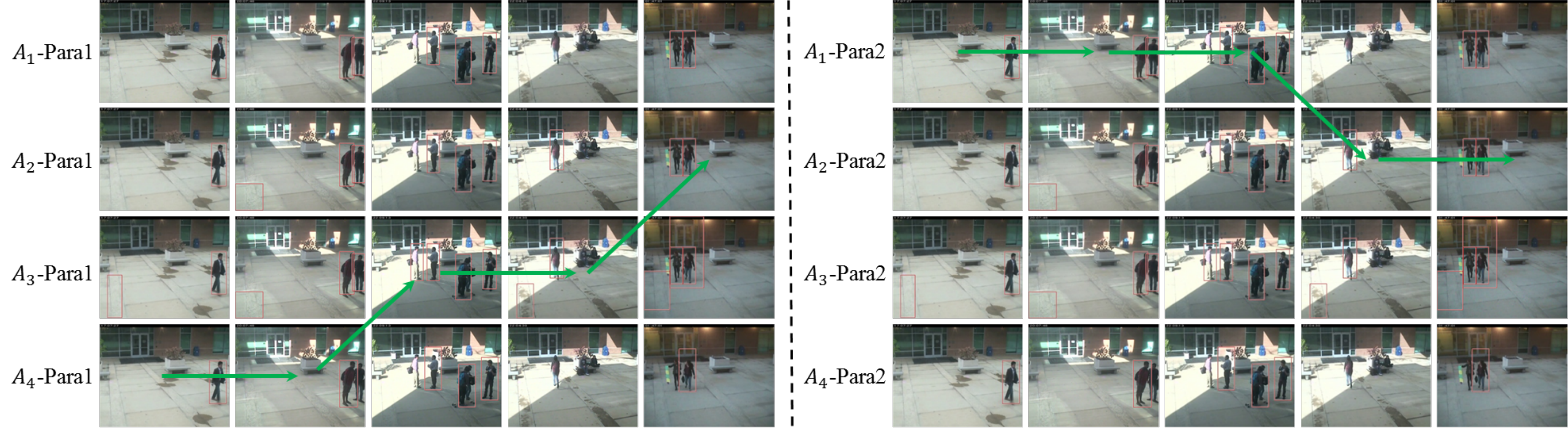}
\end{center}
\caption{Illustrations of pedestrian detection results by four algorithms $\{A_1, A_2, A_3, A_4\}$ with the parameter frames per second (FPS) in a video sequence within a day. The two parts represent two computation platforms, each of which leads to different FPSs for these four algorithms. In each part, a row denotes results of an algorithm-parameter , and each column represents an image frame. An example of the adaptive algorithm-parameter selection under a platform is shown with green arrows in the left part, which is $A_4 \rightarrow A_4 \rightarrow A_3 \rightarrow A_3 \rightarrow A_2$. In the right part, with a different platform, the algorithm-parameter selection results change to  $A_1 \rightarrow A_1 \rightarrow A_1 \rightarrow A_2 \rightarrow A_2$.}
\label{fig:SelAlgoMotivationExample}
\end{figure*}

The goal of this paper is the following. Given a set of existing computer vision algorithms and its parameters, \emph{i.e.}, algorithm-parameter combinations, for a certain problem, can we select the proper platform under certain computation and performance constraints? Also, can we automatically select the ``best'' algorithm-parameter combination on a given platform for a particular dataset (at design time this would be a benchmark; at runtime this would be real-time input video)? The answer, in most cases, will not lie in one specific algorithm-parameter combination but on an \emph{ adaptive mechanism for selecting among the set of algorithm-parameter combinations}, since the conditions in the video will likely change over time. Conditions that could trigger the switch include the lighting in the video, the number of targets in the scene, the resolution of the targets, and so on - factors which are known to affect the performance of vision algorithms. 
In the experiments, we specifically focus on the problems of pedestrian detection and tracking, since pedestrian detection is a fundamental low-level task that is crucial to higher-level tasks, \emph{e.g.} tracking, and these two tasks are known to be sensitive to environmental factors. The proposed methodology could be generally applied to other computer vision applications, with specific features selected for those applications.

An illustration of such an algorithm selection process is shown in Fig. \ref{fig:SelAlgoMotivationExample}, where the results of four pedestrian detectors are affected by the frequently-changing scales of objects, number of objects, and illumination condition.
There are two different sets of algorithm-parameter combination selection results which are obtained under two platforms. In both left and right parts of Fig. \ref{fig:SelAlgoMotivationExample}, each row shows representative image frames from a video recorded at different times of a day, and each column denotes the person detection results by four different algorithms. It is noted that each pedestrian detector achieves desired results on some image frames while does not perform well on the others. For instance, in the first frame, the detectors with the best performances are detector 1, 2 and 4, while in the second frame, the detector with the best performances changes to the first detector. In the third frame, only detector 3 successfully detects all the pedestrians. In the fourth and fifth frames, the detectors with the best performances do not lie on the same detector. The image features of the first two image frames are similar to each other. However, in the third frame, the illumination condition changes and the number of pedestrians increases. It is shown that detector 3 achieves the best performance in the third frame. Similar observations can be noticed in the rest of the images. An example of an ideal detector, which is obtained by switching between the four original detectors, is shown with green arrows in the left part. It shows the importance of developing an adaptive switching mechanism between the algorithm-parameter combinations that minimizes the detection error for each scenario. The selection of platform and algorithm-parameter combination for the left part is based on the performance constraint. In the right part, with a different performance constraint and other constraints (\emph{e.g.} energy consumption), the platform and algorithm-parameter combination is different from that of the left part. Although algorithm 3 and 4 perform well under the first platform,  their computation time under the second platform is too high and thus can not be used in this application. Such sacrifice on the performance is very necessary for many real applications when there is a need to find a balance between the performance and computational demand of algorithms.

\subsection{Overview and Contributions}

Motivated by observations from Fig. \ref{fig:SelAlgoMotivationExample}, we propose a switching algorithm which adaptively selects the best available algorithm-parameter combination along with a platform for each scenario based on the characteristics of the video under certain constraints. Our input consists of a set of existing algorithms that are well-known in the community for the specific vision task, in this case, pedestrian detection and tracking, applications we focus on in this paper. These algorithms' parameters are also known. In addition, we have datasets on which these algorithms have been tested. 
Each available algorithm has image frames as inputs and performance results as outputs. 

There are two operating phases in our proposed framework: the \emph{design time} and the \emph{run time}. At design time (offline), we learn the mapping between each unique scenario in the training data and algorithm-parameter combinations for each platform, and select the platform based on the training dataset, performance and cost constraints. The parameters that we use include frames per second (FPS) and image resolutions. The algorithm-parameter combination that obtains the best performance under the platform is then labeled as the best algorithm-parameter combination for this training scenario.

At runtime, we adapt the algorithm-parameter combination based on the real-time input, performance and cost constraints. Specifically, we segment every video sequence in the test dataset into time windows. The goal of the proposed algorithm selection process is to choose the ``best'' algorithm-parameter combination for each video segment. This is done by two steps. The first step is to compute a similarity function between the test video segment and all the training scenarios over a learned manifold of image features shared by the training and test dataset. This method has been referred to as domain adaptation in the literature~\cite{Gong12,Gopalan11}. 
The output is the training scenario that the test video is closest to in this space. In the second step, the ``best'' algorithm-parameter combination for a video segment is obtained by selecting the algorithm-parameter combination with best performances in the selected training scenario. 
Note that in principle, we can also adjust the platform (to some degree, depending on the platform capability), but this is out of the scope of this paper.

We demonstrate the efficacy of the proposed approach on multiple well-known datasets. We apply  10 algorithm-parameter combinations on 3 public datasets \cite{Ess08,Wojek09,PETS09Dataset}. We show how to choose the ``best'' algorithm-parameter combination for each time window of image frames  through switches from one algorithm-parameter combination to another in a dataset under a performance constraint.
It is proved that the proposed approach is able to obtain the optimal or close-to-the-optimal performances among all the algorithm-parameter combinations' performances given certain performance constraints.

\begin{figure*}
\begin{center}

\includegraphics[width=1\linewidth]{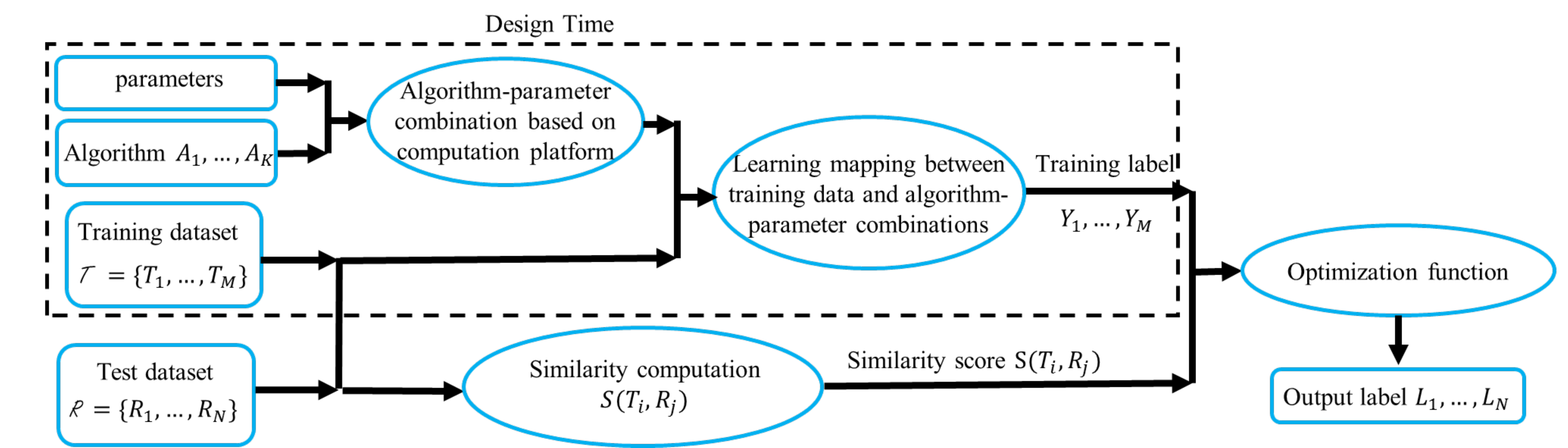}
\end{center}
\caption{Overall Methodology. The algorithms $\mathcal{A}=\{ A_1, \cdots, A_K \}$ are combined with a couple of parameters to generate the algorithm-parameter combinations under certain performance constraints. We learn the mapping between the training data and each algorithm-parameter combination, and obtain the training label $\mathcal{Y}=\{Y_1, \cdots, Y_K \}$. The feature similarity scores between the training and test datasets are calculated by $\mathcal{T}$ and $\mathcal{R}$. A cost function with two steps is defined and solved in Sec. \ref{Sec:SelAlgoAdaptiveAlgo}.  }
\label{fig:SelAlgoOverall}
\end{figure*}

\subsection{Related Works} \label{Sec:RelatedWorks}

Algorithm selection has been studied in recent years in a few works. In~\cite{Yong05}, image segmentation algorithms are selected on different images. Features are learned by support vector machine (SVM) and the performance of each algorithm is mapped to a four-bin ranking vector based on the correlation between features. The results are shown to be effective on 1000 synthetic images. In~\cite{Aodha10}, the goal is to segment pixels in an image into different regions that are suitable to different algorithms. Different features are classified by a random forest classifier, and different optical flow algorithms are automatically selected.

Our work is different from these two approaches. We consider the problem of automatically switching the algorithm based on the scene similarity between a test time window and all the unique scenarios in the training dataset. Our proposed algorithm does not learn which specific feature to be used for a dataset, and does not need manual analysis of the feature-performance correspondence. This is more general than~\cite{Aodha10}, where the effects of the features on the training dataset are manually analyzed to obtain the correlations between features and algorithms, \emph{e.g.} which feature has an impact on a specific data.  The methodology of domain adaptation that we use finds the underlying correspondences between features while the approaches\cite{Yong05,Aodha10} do not investigate this issue.  In~\cite{JosephWang14}, budget constraints are taken into consideration as the leverage rule between different algorithms in the context of handwriting recognition and scene categorization. The algorithms are selected based on a binary tree. Our work does not only consider the budget constraints. Instead, our proposed framework considers both budget constraints and the performances of each algorithm. We also investigate the possibility to change computation parameters to improve the algorithm performances.

To the best of our knowledge, this is the first work that adaptively selects algorithms with applications on pedestrian detection and tracking. We briefly introduce some related works on these two applications. The most widely used pedestrian detector in the past decade is the Histograms of Oriented Gradients (HOG) detector~\cite{Dalal05}, where the HOG feature is developed and a linear SVM is adopted to classify HOG features.  The part-based model (PBM) that is developed in~\cite{Felzenszwalb10,Felzenszwalb08} applies HOG features on a part-based multi-component model and achieves very good performance on some datasets. The work in~\cite{Gkioxari14} considers a deformable part model with k parts as k-poselets, and uses a separate HOG template to model the appearance. A summary of the works of person detection can be found in~\cite{Enzweiler09}. The work in~\cite{Zeng14} looks into the problem of how a previously trained classifier can be adapted to a target dataset and proposes deep networks which jointly solve the problem of target detection as well as reconstruction of the target scene. Our method differs from this in two important ways. First, we build upon existing well-known algorithms whose performance is very well understood and choose the best algorithm for each video segment. Second, the method proposed here can be applied online as new test data is available.

In the area of multi-target tracking, most state-of-the-art works focus on solving the problem of data association, given that the detections are available. The works \cite{Song10,YangBo12} adopted the bipartite graph matching method to find out initial detection association results, and used the statistics or other properties of the associated tracks to obtain the final tracking results. The works \cite{Qin12,ZhangShu13,Chen14} developed complex detection association models based on the grouping behaviors between targets. In the works \cite{Pirsiavash11,Butt13,DehghanPaper215}, the problem of multi-target tracking was modeled as a network-flow problem. Although the state-of-the-art results have been obtained recently, the computations of these algorithm are usually too high to be adopted in the applications that require certain budget constraints or processing speed. For example, the works \cite{ZhangShu13,DehghanPaper115} have complex graph structures which make the learning and inference of the graphs time consuming. To meet the requirement of the computation time, we adopt a simple yet effective approach that has been widely used as the baseline algrotihm in the works \cite{Song10,YangBo12,Qin12,ZhangShu13,Chen14}. The details are provided in the experimental section.

\section{Methodology}

\subsection{Problem Description}

We assume the availability of a number of algorithms for the problem. Representative parameters of these algorithms are also known. Our goal is to answer the following questions: for every part of an unknown dataset, is it possible to automatically select an algorithm-parameter combination along with a platform among all available algorithm-parameter combinations that achieves the best result under certain performance constraints? And for the entire unknown dataset, what is the best strategy to switch between algorithms?

In our problem, the input is the set of $K$ available  algorithms $\mathcal{A}=\{ A_1, \cdots, A_K \}$ with different parameters and the dataset on which they are evaluated. We call this the training dataset $\mathcal{T}= \{ T_1, \cdots, T_M \}$, where $T_i$ represents the $i$-th unique scenario in  $\mathcal{T}$. The segmentation of $\mathcal{T}$ can be done by any data classification methodology. We combine algorithms with different parameters to obtain a set of algorithm-parameter combinations, denoted by $\mathcal{B}=\{ B_1, \cdots, B_H \}$.

 Under each platform in the design time, we apply every algorithm-parameter combination $B_h$ in $\mathcal{B}$ on each $T_i$ in $\mathcal{T}$. Given a constraint such as a performance constraint, we select a platform with the corresponding computation capability. We then select the algorithm-parameter combination that performs the best as the training label $Y_i$ under the constraint.

The unknown dataset is called the test dataset $\mathcal{R}$. In $\mathcal{R}$, we assume that there are totally $N$ time windows. Every time window of images is denoted as $R_j, j = 1, \cdots, N$. In the runtime, the selection of algorithms for $R_j$ is represented by $L_j$.  Given the pairs $(T_i, Y_i)$ under the same performance constraint, the problem is how to find the unknown label $L_j$ for each $R_j$ that is in $\mathcal{R}$. All the notations are highlighted in Table \ref{tab:SelAlgoNotation}.

\begin{table*}[htbp]

\begin{center}
% \footnotesize \addtolength{\tabcolsep}{.3pt}

\begin{tabular}{|c|c|}
\hline
$\mathcal{A}$    & the set of available algorithms $A_1, \cdots, A_K$ \\
\hline
$\mathcal{B}$    & the set of available algorithm-parameter combinations $B_1, \cdots, B_H$ \\
\hline
$\mathcal{T}$   & training dataset $T_1, \cdots, T_M$  \\
\hline
$\mathcal{R}$   & test dataset $R_1, \cdots, R_N$  \\
\hline
 $Y_i$ & the label of the training data $T_i \in \mathcal{T}$  \\
\hline
 $L_j$ & the label of the test data $R_j \in \mathcal{R}$   \\
\hline
 $t_i$ & the feature of $T_i$, the dimension of which is $a$   \\
\hline
 $r_j$ & the feature of $R_j$, the dimension of which is $a$   \\
\hline
$b$ & the dimension of the subspace of $t_i$ and $r_j$   \\
\hline
 $x_i$ & the basis of subspace of $t_i$, the dimension of which is $a \times b$  \\
 \hline
 $z_j$ & the basis of subspace of $r_j$, the dimension of which is $a \times b$   \\
 \hline
 $\tilde{x}_i$ & orthogonal to $x_i$, the dimension of which is $a \times (a-b)$  \\
 \hline
 $\tilde{z}_j$ & orthogonal to $z_j$, the dimension of which is $a \times (a-b)$   \\
\hline
$W_{ij}$ & geodesic kernel   \\
\hline
$\theta (y)$ & geodesic flow parametered by $y$ in Eq. \ref{Eq:SelAlgotheta}  \\
\hline
$\Lambda_i$ & diagonal matrices in Eq. \ref{Eq:SelAlgoW} \\

\hline
\end{tabular}
\end{center}

\caption{Notation Table.}
\label{tab:SelAlgoNotation}
\end{table*}

\subsection{Solution Overview}

The overview of our solution is shown in Fig. \ref{fig:SelAlgoOverall}. In an unknown test dataset $\mathcal{R}$, every video sequence/image set is segmented into a sequence of non-overlapping time windows $R_j$. The output of the algorithm, the label set $\mathcal{L} = \{ L_1, \cdots, L_N \}$, is obtained by a two-step cost function. This cost function is able to automatically select the best algorithm-parameter combination on a specific time window $R_j$ under certain platform which is determined by the performance constraint. In the first step, we measure a similarity score $S(T_i,R_j)$ between $R_j$ and every training scenario $T_i$ in the training dataset $\mathcal{T}$, and find the scenario $T_{i^*}$ that is most similar to $R_j$, \emph{i.e.}, with largest $S(T_{i^*}, R_j)$. In the second step, we find the ``best'' algorithm-parameter selection for $T_{i^*}$, and choose that selection for $R_j$.

The structure of our solution is as follows. In Sec. \ref{Sec:SelAlgoPerfomanceTestData}, we introduce how to calculate the similarity score $S(T_i,R_j)$ between the time window $R_j$ in the test dataset and a particular scenario $T_i$ in the training dataset. In Sec. \ref{Sec:SelAlgoAdaptiveAlgo}, we introduce the overall two-step cost function that is able to find the best algorithm-parameter selection for every time window in $\mathcal{R}$.

\subsection{Similarity Scores between Training Scenarios and Test Time Windows}
\label{Sec:SelAlgoPerfomanceTestData}

Following the similarity definition in \cite{Song10}, the similarity between $R_j$ and $T_i$, denoted by $S(T_i,R_j) $,  is calculated as an exponential function of feature distance:
\begin{equation}\label{Eq:SelAlgoSimScore}
S(T_i,R_j) = e^{-d(T_i,R_j)}
\end{equation}
where  $d(T_i,R_j)$ represents the feature distance between $T_i$ and $R_j$, whose computation is shown in below. 

\subsubsection{Feature Distance Computation}
\label{Sec:FeatureDist}

\begin{figure}
\begin{center}

\includegraphics[width=1\linewidth]{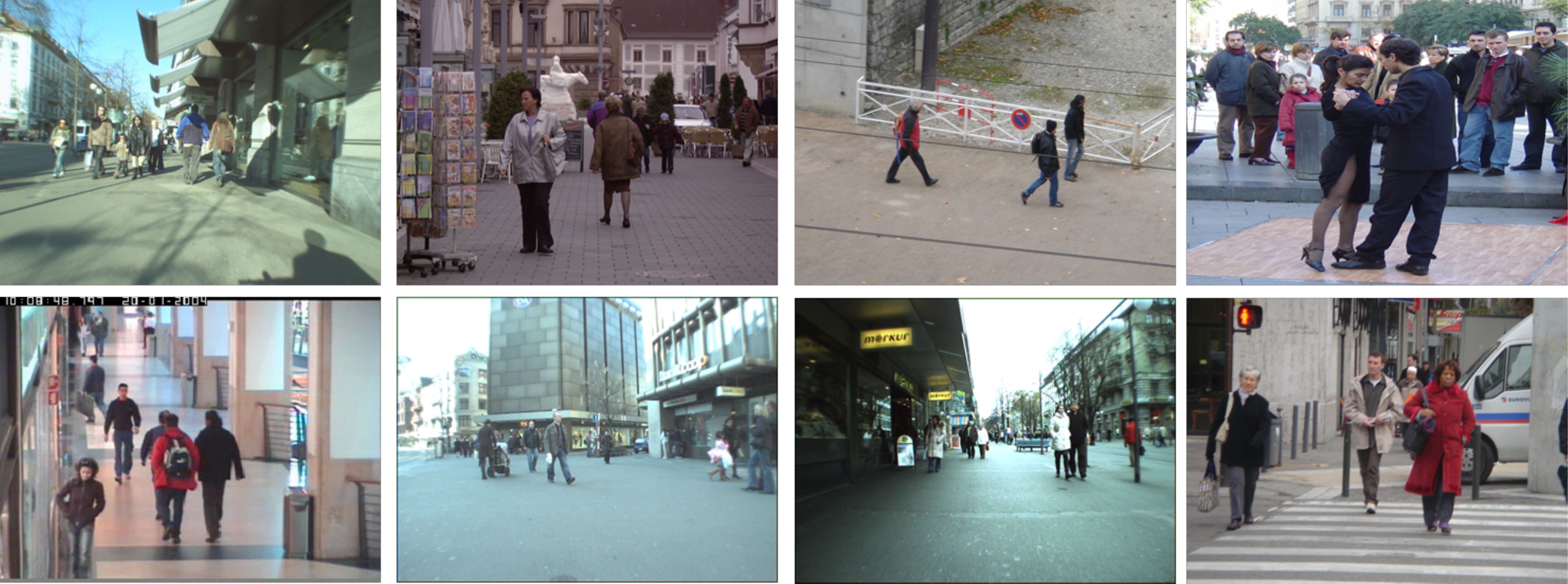}
\end{center}
\caption{Examples of mismatches between feature distributions. Every column of images do not share the same feature space. The application of domain adaption indicates that the same pedestrian detector should be applied to both rows of each column. }
\label{fig:SelAlgoDomainAdap}
\end{figure}

In this section, we provide a solution to $d(T_i,R_j)$ in Eq. \ref{Eq:SelAlgoSimScore}. Different from  \cite{Aodha10}, where feature distances are directly computed, we consider the mismatch between the training data and the test data. This mismatch can come from many sources, \emph{e.g.}, pose, illumination, image quality, etc. In other words, even though the training and test data have the features lying in different spaces, a domain shift might indicate similar distributions of the two sets of features. An example is shown in Fig. \ref{fig:SelAlgoDomainAdap}, where each column of images does not have the same feature distributions in terms of illumination, size of pedestrians and etc. However, the pedestrian detection experiments show that the same algorithm should be applied to each column of images to achieve the best performance. It represents that directly calculating the feature distance between two data may mislead to wrong classification results. It is highly likely that there is an underlying space that is shared by features of both training and test data. If the features of $T_i$ and $R_j$  share similar distributions on such a space, there is a high chance that the same algorithm can be applied to both $T_i$ and $R_j$. Finding such a space is often known as the problem of domain adaptation. Our solution is motivated by the approaches in \cite{Gong12,Gopalan11}, where the mapping between the training data and the test data is modeled as geodesic flow.

 The key idea of domain adaptation is to project both the training data and each video segment of the test data into subspaces to learn domain-invariant features. A challenge is how to determine and select the subspace that is shared by both training data and test data. We use a geodesic flow curve to link the training and test data on a Grassmann manifold (similar to\cite{Gong12,Gopalan11}), which is a special type of Riemannian manifold \cite{Harandi13}. As long as the projections of the two datasets on the Grassman manifold are similar, the features of the data also share similar distributions. Now the problem is converted to the computation of geodesic flow, as explained below. 

Denote  the features of $T_i$ and $R_j$ as $t_i  \in \mathbb{R}^a$ and $r_j \in \mathbb{R}^a$ individually. We denote $b$ as the dimension of the subspace of $t_i$ and $r_j$. Performing principal component analysis (PCA) on $t_i$ and $r_j$, we can obtain $x_i \in \mathbb{R}^{a \times b}$ and $z_j \in \mathbb{R}^{a \times b}$ which are the basis vectors for the subspaces  of $t_i$ and $r_j$.  The orthogonal matrix of $x_i$ is defined as $\tilde{x}_i \in \mathbb{R}^{a \times (a-b)}$, and that of  $z_j$ is defined as $\tilde{z}_j \in \mathbb{R}^{a \times (a-b)}$.

The work in \cite{Gong12} provides a closed loop solution to the geodesic flow between $t_i$ and $r_j$,  which is

\begin{equation}
\label{Eq: SelAlgoCloseloop}
 t_i^T W_{ij} r_j= \int_{0}^{1} {(\theta(y)t_i)}^T {(\theta(y)r_j)} dy.
\end{equation}
where $\theta(y)$ is a constructed geodesic flow function parameterized by a continuous variable $y \in \left[0, 1 \right]$ and $W_{ij}$ is the kernel function that is defined below.

The term $\theta(y)$ in Eq. \ref{Eq: SelAlgoCloseloop} represents how to smoothly project a feature $t_i$ into $r_j$, where $\theta(y) t_i$ projects a feature into the $y$-th subspace on the Grassmann manifold. $\theta(y)$ is defined as

\begin{equation} \label{Eq:SelAlgotheta}
\theta(y) =\left\{
\begin{array}{cc}
x_i,& {\rm if }\  y=0,\\
z_j,&{\rm if}\ y=1,\\
x_i U \Sigma_1(y) - \tilde{x}_i V \Sigma_2(y), &{\rm otherwise}

\end{array}
\right.
\end{equation}
where $U$, $V$, $\Sigma_1$ and $\Sigma_2$ are obtained by singular value decomposition (SVD) of $x_i^T z_j$ and $\tilde{x}_i^T z_j$. The index $y$ denotes the $y$-th subspace. In other words, $y$ is a continuous variable and $\theta(y)$ parameterizes infinite number of $y$ to construct geodesic flow.

Looking back to Eq. \ref{Eq: SelAlgoCloseloop}, we find that it calculates the geodesic flow over all the $y$, which means that the original features with projections are expanded to all subspaces. In this case, $W_{ij}$ is of importance since it induces inner projects between features with infinite dimensions.  In theory, $W_{ij}$ is the kernel between $t_i$ and $r_j$ and can be calculated by 

\begin{equation} \label{Eq:SelAlgoW}
W_{ij} = \begin{bmatrix}
x_i U & \tilde{x}_i V
\end{bmatrix}
\begin{bmatrix}
\Lambda_1 & \Lambda_2 \\
\Lambda_2 & \Lambda_3
\end{bmatrix}
\begin{bmatrix}
U^T x_i^T \\
V^T \tilde{x}_i^T
\end{bmatrix},
\end{equation}
where the matrices $\Lambda_1$ to $\Lambda_3$ are diagonal matrices. The elements of $\Lambda_1$ to $\Lambda_3$ come from $\Sigma_1$ and $\Sigma_2$ in Eq. \ref{Eq:SelAlgotheta}. The details of the derivation can be found in \cite{Gong12}.
 
In summary, calculating the feature distance kernel assumes that the subspaces of $t_i$ and $r_j$ lie on a Grassmann manifold. Eq. \ref{Eq: SelAlgoCloseloop} constructs geodesic flow between $t_i$ and $r_j$, where the correlations between $t_i$ and $r_j$ are parameterized by the continues variable $y$. The projection of the feature $t_i$ on the Grassmann manifold is $\theta(y) t_i$, and that of the feature $r_j$ is $\theta(y) r_j$. The kernel inner product of $t_i$ and $r_j$ essentially represents how close their subspace projections are.

The feature distance $d$ in Eq. \ref{Eq:SelAlgoSimScore} can be calculated using kernel distance \cite{Phillips11} given the calculated kernel $W_{ij}$. The kernel distance is able to calculate the distance between two sets of points which lie on geometric surfaces, i.e., the manifold that $t_i$ and $r_j$ lies on. The kernel distance  between $T_i$ and $R_j$ is defined as in \cite{Phillips11}

\begin{equation} \label{Eq:SelAlgodistance}
d(T_i,R_j) = t_i^T W_{ij} t_i + r_j^T W_{ij} r_j - 2 t_i^T W_{ij} r_j.
\end{equation}

In Eq. \ref{Eq:SelAlgodistance}, the first two terms are self-similarities between the feature $t_i$ of the training data $T_i$ and the feature $r_j$ of the test data $R_j$ individually. The third term, that is defined in Eq. \ref{Eq: SelAlgoCloseloop}, is the inner product of the two features that measures how close they are correlated to each other.

\subsection{Adaptive Algorithm-Parameter Selection Cost Function}
\label{Sec:SelAlgoAdaptiveAlgo}

The ultimate goal of our proposed approach is to automatically select an algorithm-parameter for a time window of images $R_j$ in the test data under performance and cost constraints. We formulate the selection process in the test dataset $\mathcal{R}$ as a two-step optimization function.

\subsubsection{Step 1 of the Cost Function}

We obtain the training scenario that is closest to the test time window $R_j$. This training scenario $T_{i^*}$ is obtained by finding the maximum similarity between all the training scenarios and $R_j$ 

\begin{equation} \label{Eq:SelAlgoCostFn1}
T_{i^*}  = \max_{i}  S(\{ T_1, \cdots, T_i, \cdots, T_M \},R_j),
\end{equation}
where $S$ denotes the similarity function that is defined in Eq. \ref{Eq:SelAlgoSimScore}.

\subsubsection{Step 2 of the Cost Function}

In the design time, given constraints such as system performance and cost, we select a platform configuration $C$ that could meet the requirements. In the run time, we will find the algorithm-parameter combination that performance the best for $T_{i^*}$ (obtained in the first step of the cost function) given the selected platform configuration $C$, and choose this algorithm-parameter combination for $R_j$. 

The algorithm-parameter selection, which is essentially the output label $L_j$, is obtained by selecting the best performance $P$ of the training scenario $T_{i^*}$ under the selected platform configuration $C$ 

\begin{equation} \label{Eq:SelAlgoCostFn2}
L_j = \max_{h} P(B_h|T_{i^*}, C ),
\end{equation}
where the superscript $h$ denotes the $h$-th algorithm-parameter combination. 

Note that in the design time, we exhaustively apply every algorithm-parameter combination $B_h$ on each training scenario $T_i$. The algorithm-parameter combination obtaining the minimum error (best performance), which was calculated in the design time, is selected as the solution to Eq. \ref{Eq:SelAlgoCostFn2} in the run time. The details of computation parameter selections are shown in Sec. \ref{Sec:SelAlgoExpSetup}.

\section{Experiments}

\subsection{Experimental Setup} \label{Sec:SelAlgoExpSetup}

In the experiments, we show results of our method on two applications: pedestrian detection and pedestrian tracking. There are 5 state-of-the-art available detection algorithms: HOG \cite{Dalal05}, PartBased \cite{Felzenszwalb10}, Cascades \cite{Cevikalp12}, ACF \cite{Dollar14}, and LDCF \cite{Nam14}. The public datasets that are used as the test datasets are:  INRIA \cite{Dalal05}, ETHMS \cite{Ess08}, and TUD Stadtmitte \cite{Andriluka10}.

In the application of pedestrian tracking, we adopt the baseline algorithm shared by  \cite{Song10,YangBo12,Qin12,ZhangShu13,Chen14}, due to its computation efficiency. This algorithm is based on the detection association methodology. Thus the effects of different detection results on tracking can be demonstrated by using the same tracking module.  Any other detection association algorithm can be adopted, as long as the computation requirements are satisfied. Among the detection datasets, we use all the datasets expect for the INRIA dataset and the TUD-Brussels dataset for tracking. The reason is that the INRIA dataset is not composed of consecutive image frames and thus is not suitable for the problem of tracking, and that the TUD-Brussels dataset was recorded by a fast-moving platform, which makes every pedestrian only exist 1-2 frames.

We extract four different features that are used for distance calculation in the experiments: HOG features \cite{Dalal05}, SIFT features \cite{Lowe99}, gradient features \cite{Oliva01}, and texture features \cite{GLCMFeature}. We resize every image frame to $64 \times 128$ and use the methodology of Principle Component Analysis (PCA) to reduce the dimension of the feature combinations to 1288, where the HOG features have the size of 800 due to its dominance in pedestrian detection.

In the training dataset $\mathcal{T}$, all the scenarios are clustered into 15 unique scenarios. The number of unique scenarios is determined based on the observation of the characteristics of the training data, \emph{e.g.} the lighting condition, the density of the scenarios and etc.  In the test dataset $\mathcal{R}$, all the videos/image frames are segmented into different time windows. The length of a time window is set to be 30 frames except for the INRIA dataset. The reason is that the INRIA dataset is composed of non-consecutive image frames, and we set the length of a time window to be 10.  The computation complexity of our algorithm is low, since the cost function only needs two max operations. Typically the computation time for each time window is less than 0.5s on an Intel i7 platform.

\subsection{Results of Algorithm-Parameter Selection on Various Platforms}

We investigate the effects of algorithm-parameter combination switches in every dataset.  In the design time, we estimate the classification threshold of each detection algorithm that leads to FPPI $=1$. In the test dataset, we keep the detections with the scores greater than this threshold for each algorithm.  We evaluate the detections for each time window, where the number of missed detections is used as the evaluation metrics. The reason of using missed detections as evaluation metrics is that the overall FPPI of every algorithm is fixed to be around 1, and the number of missed detections is assumed to be dominant in determining the performance for each time window of the test dataset.  In the problem of tracking, we adopt four evaluation metrics: mostly tracked (MT), mostly lost (ML), number of ID switches (IDS) and false positive (FP).

\begin{figure*}
\centering       
\begin{subfigure}[b]{0.45\textwidth}
\includegraphics[width=0.9\textwidth]{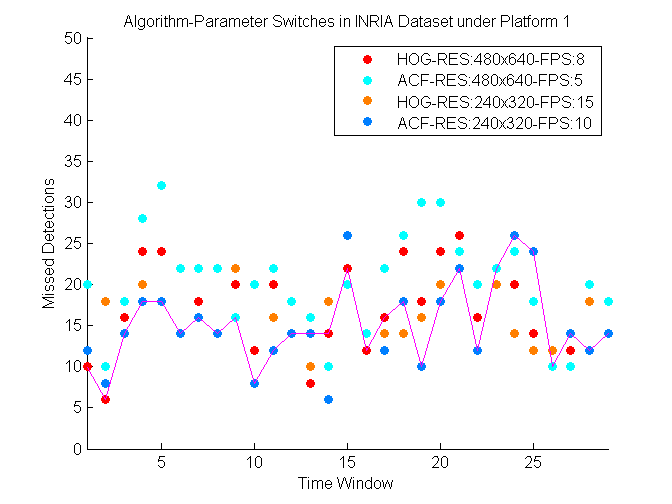}
\end{subfigure} 
\begin{subfigure}[b]{0.45\textwidth}
\includegraphics[width=0.9\textwidth]{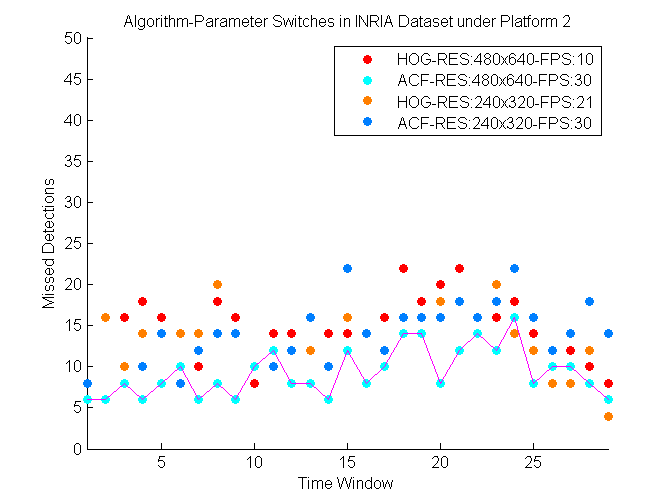} 
\end{subfigure}   
\caption{ Algorithm-parameter selection results with the application of pedestrian detection on INRIA dataset with four algorithm-parameter combinations under two platforms. In each subfigure, x-axis represents the time window and y-axis represents the number of missed detections. Each subfigure shows the results under a platform. It is shown that our algorithm-parameter selection process can select the low-error results in most time windows under both platforms. For instance, in (a), the selected algorithm-parameter only fails to select the best performance at the time window 13, 14, 17, 24, 25, and 27 among totally 29 time windows. The selected algorithm-parameter switches between  HOG-RES:480x640-FPS:8 and ACF-RES:240x320-FPS:10, each of which obtains the best results on some time windows. ACF-RES:240x320-FPS:10 obtains the best results in most time windows under the platform 1.  Given a stricter performance constraint, the platform 2 is selected in (b).  The algorithm ACF-RES:480x640 performs the best in most time windows. It is because the high performance requirement leads to a powerful platform selection, which is easily to process high FPS and resolutions. The selected algorithm follows the correct trend, and lies in the ACF-RES:480x640 in all time windows. It is shown that the select algorithm does not obtain the best performance only for four time windows.  }
 \label{fig:SelAlgoSwitch1}
\end{figure*}

We show the validation and importance of the algorithm selection process in Fig. \ref{fig:SelAlgoSwitch1}, where the numbers of missed detections of the INRIA dataset are shown under two platforms. 4 algorithm-parameter combinations, whose FPS are shown in the legends, are used under each platform. In each subfigure, x-axis represents time windows and y-axis represents the number of missed detections. The selected algorithms are shown by pink curves. Overall, our approach selects the best algorithm-parameter combinations in most time windows for both datasets under both platforms.  It is also noted that the selected algorithms are not the same under different platforms. For instance, ACF-240x320 performs well in most time windows with FPS=10  under the first platform, while ACF-480x640 obtains the best results in most time windows with FPS=30 under the second platform. Our algorithm selection process successfully captures such algorithm changes within a platform and between platforms. The results of (b) are better than that of (a) because the designed platform is more powerful than that of (a), and thus makes the FPS of each algorithm higher than that of (a). A detailed description of Fig. \ref{fig:SelAlgoSwitch1} is shown in the figure caption.

\begin{figure*}
\centering       
\begin{subfigure}[b]{0.45\textwidth}
\includegraphics[width=0.9\textwidth]{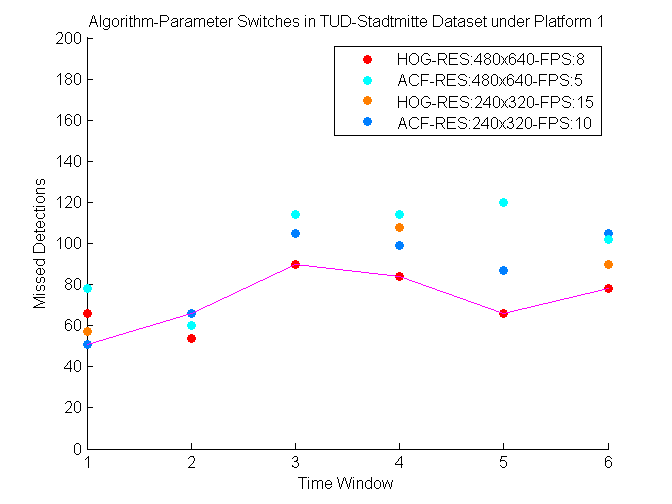}
\end{subfigure}
\begin{subfigure}[b]{0.45\textwidth}
\includegraphics[width=0.9\textwidth]{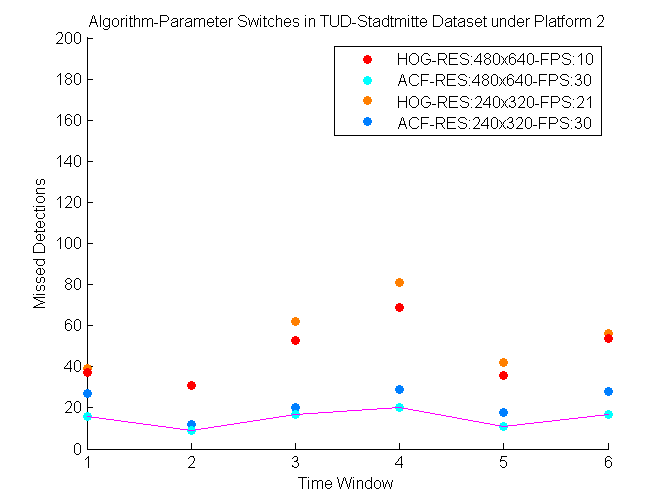}
\end{subfigure} 
\begin{subfigure}[b]{0.45\textwidth}
\includegraphics[width=0.9\textwidth]{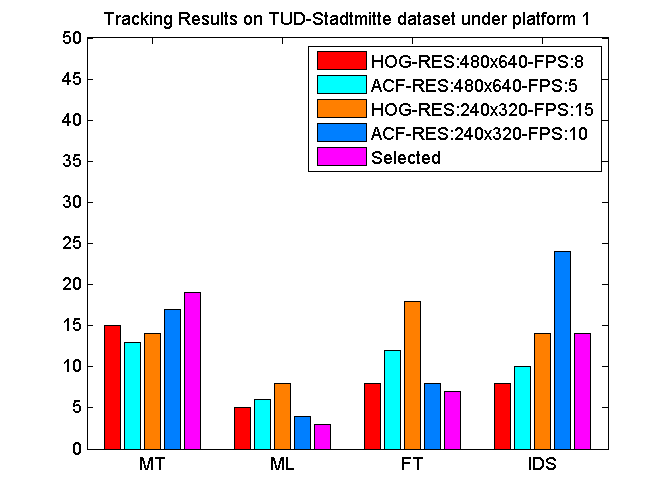}
\end{subfigure} 
\begin{subfigure}[b]{0.45\textwidth}
\includegraphics[width=0.9\textwidth]{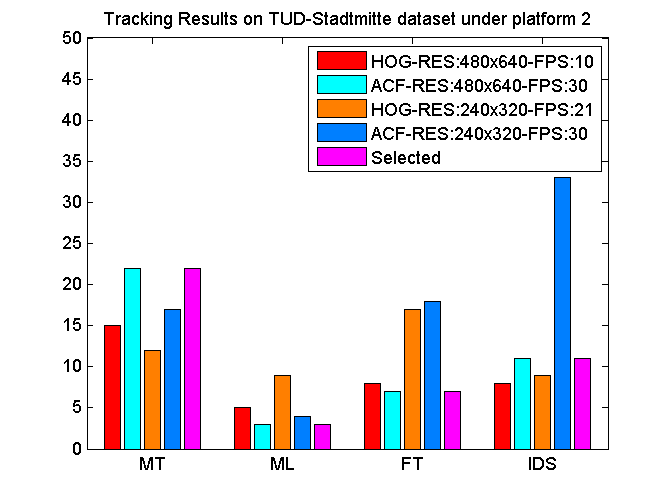}
\end{subfigure}
\caption{ Algorithm-parameter selection results on TUD-Stadtmitte dataset with four algorithm-parameter combinations under two platforms. In the top subfigures, x-axis represents the time window and y-axis represents the number of missed detections. The left subfigures show the results under the first platform and the right subfigures show the results under the second platform. The top subfigures demonstrate the detection results under different platforms given performance constraints.  In (a), the selected algorithm-parameter only fails to select the best algorithm-parameter at the second time window.  Given a new performance constraint, the platform is chosen as (b), where the selected algorithm-parameter does not lie in ACF-RES:480x640 as the first platform. (c) and (d) show the tracking results. In (c), the selected algorithm-parameter obtains better MT, ML and FT than any single algorithm-parameter.  Though its IDS is a little higher than other two algorithms, its overall performance is the best. In (d), the tracking performance of the selected algorithm is the same as that of ACF-RES:480x640, which obtains the best performance for all time windows.}
 \label{fig:SelAlgoSwitch2}
\end{figure*}

\begin{figure*}
\centering       
\begin{subfigure}[b]{0.45\textwidth}
\includegraphics[width=0.9\textwidth]{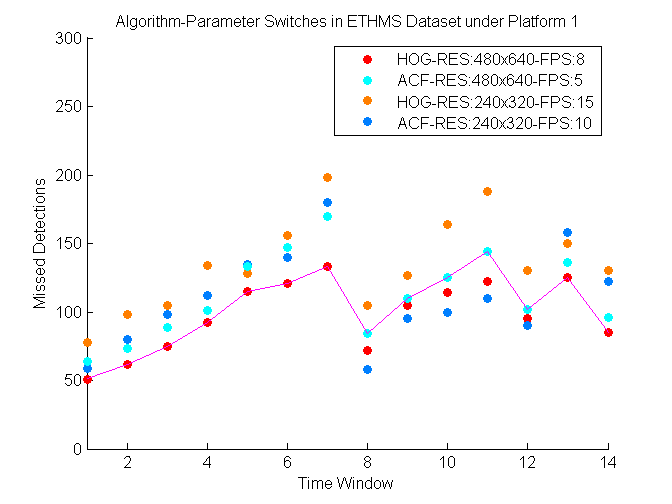}
\end{subfigure}
\begin{subfigure}[b]{0.45\textwidth}
\includegraphics[width=0.9\textwidth]{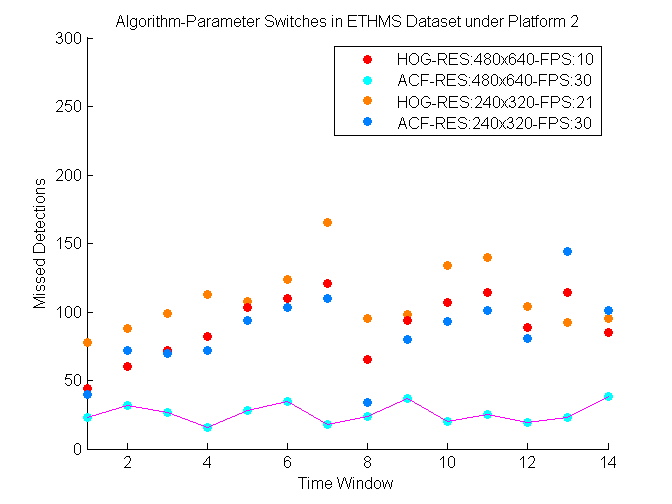}
\end{subfigure}
\begin{subfigure}[b]{0.45\textwidth}
\includegraphics[width=0.9\textwidth]{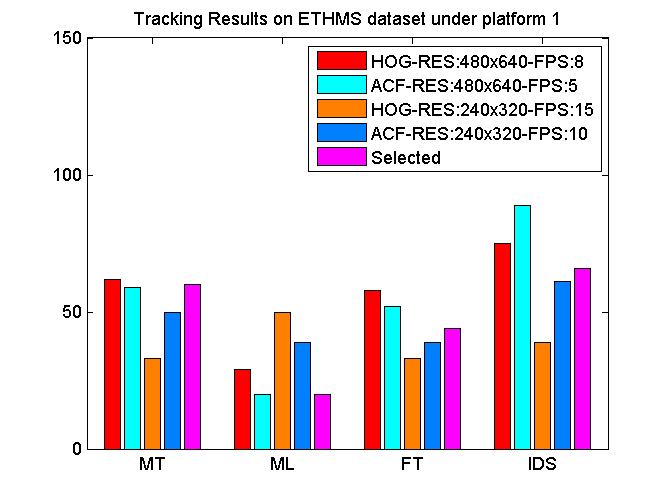} 
\end{subfigure}
\begin{subfigure}[b]{0.45\textwidth}
\includegraphics[width=0.9\textwidth]{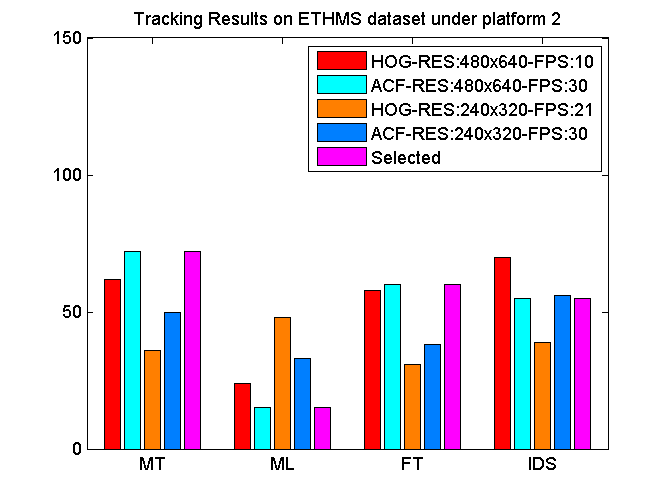}
\end{subfigure}
\caption{ Algorithm-parameter selection results on ETHMS dataset with four algorithm-parameter combinations under two platforms. Note that we only show parts of the time windows of ETHMS dataset to clearly denote how the results switch. In the top subfigures, x-axis represents the time window and y-axis represents the number of missed detections. The left subfigures show the results under the first platform and the right subfigures show the results under the second platform given a new performance constraint. The top subfigures demonstrate the detection results under different platforms.  The selected algorithm-parameter can capture the lowest detection errors in most time windows under the two platforms given different performance constraints. Different from Fig. \ref{fig:SelAlgoSwitch1} and Fig. \ref{fig:SelAlgoSwitch2}, where the selected algorithm-parameter mainly switches between HOG-RES:480x640 and ACF-RES:240x320 with different FPSs under the first platform, the selected algorithm-parameter of ETHMS dataset also selects ACF-RES:480x640 with different FPSs under the first platform. In the second platform (b), the selected algorithm always lies in ACF-RES:480x640, which is the same as the results of Fig. \ref{fig:SelAlgoSwitch1} and Fig. \ref{fig:SelAlgoSwitch2}. In the tracking results of both (c) and (d), the selected algorithm-parameter can obtain the best performance, which also supports the detection algorithm selection.}
 \label{fig:SelAlgoSwitch3}
\end{figure*}

In Fig. \ref{fig:SelAlgoSwitch2}, we show both detection and tracking results on the TUD-Stadtmitte dataset, where the top two subfigures show the detection results and the bottom two show the tracking results. The detection algorithm-parameter selections are different under the two platforms, leading to different tracking results. In (a), the detection algorithm-parameters switch between ACF-RES:240x320 and HOG-RES:480x640 with the corresponding FPSs under the platform 1. In (b), the selection always lies in ACF-RES:480x640 with FPS=30 under the platform 2. It is reasonable because the performance requirement is strict in (b). Such a performance requirement leads to a powerful platform selection that allows a high resolution and FPS of an algorithm. In (c) and (d),we show the tracking results with the four evaluation metrics: MT, ML, FT and IDS. A good tracker should have high MT and low ML, FT, and IDS. In (c), the selected algorithm-parameter obtains better results than any single algorithm-parameter combination under the first platform. It is demonstrated that the tracking performances of the selected algorithm follow the trend of the detection performances. The results prove the effectiveness of adaptively selecting algorithm-parameters. In (d), the results of the selected algorithm follows the trend of (a), where ACF-RES:480x640 obtains the best result among all the algorithm-parameter combinations. Similarly, we also show results on ETHMS dataset under the same platforms in Fig. \ref{fig:SelAlgoSwitch3}. Detailed descriptions of Fig. \ref{fig:SelAlgoSwitch2} and \ref{fig:SelAlgoSwitch3} are illustrated in their captions.

\subsection{Results of Platform Selection}
 
Different performance constraints may lead to different selections of platforms and algorithm-parameter combinations.. In our experiments, we consider two parameters of each algorithm: FPS and image resolutions.  If the performance constraints (e.g., tracking accuracy) are moderate, a platform with low computation and low cost may be chosen, and the most suitable algorithm and its parameters can then be chosen accordingly. If the performance requirements are high, a platform with high computation capability may be needed, and correspondingly a \emph{different} set of algorithms and parameters may be chosen. This is the essence of selecting platform and algorithm (including parameters) based on design requirements (including performance requirement and other constraints such as energy consumption or cost).

\begin{table*} 
\centering
\caption{Algorithm-parameter combinations under two platforms.}
\label{tab:SelAlgoFPSandResol}
\begin{tabulary}{\textwidth}{| c | c | c | c | c |}
\hline
& \multicolumn{2}{c|}{\bf{Platform 1}} & \multicolumn{2}{c|}{\bf{Platform 2}}\\
\hline
      & \bf{FPS} / \bf{Resolution} & \bf{FPS} / \bf{Resolution} & \bf{FPS} / \bf{Resolution} & \bf{FPS} / \bf{Resolution}\\
\hline 
\bf{HOG} & 15 / 240x320 & 8 / 480x640 & 30 / 240x320 & 15 / 480x640 \\
\hline

\bf{ACF} & 10 / 240x320 & 5 / 480x640 & 20 / 240x320 & 10 / 480x640  \\
\hline
        
\end{tabulary}   
\end{table*}

In the experiments, we consider different performance requirements that lead to two different platform selections. Then, for each platform, we consider the set of algorithm-parameter combinations that are computationally feasible on the platform and select the one that provides the best performance for pedestrian detection. We have tested two PCs as candidate platforms. The algorithm-parameter combinations under each platform are shown in Table \ref{tab:SelAlgoFPSandResol}, where the parameters are obtained by running codes on the two platforms. In Fig. \ref{fig:SelAlgoSwitch1}, we can see that for platform 1, ACF-RES:240x320-FPS:10 provides the best performance while for platform 2, ACF-RES:480x640-FPS:30 provides the best performance.  Note that we have also tested the algorithms PartBased, Cascades, and LDCF. However, because of the low FPSs of these algorithms yielding bad performances, we only show results of HOG and ACF, which obtain reasonable results with adaptive selecting of the platform and algorithm-parameter combination under performance constraints.

\section{Conclusion}

In this paper, we present a novel approach to adaptively select the best platform and algorithm-parameter combination. Our approach consists of an offline design step and an online running step.  In the design step, we select platform configuration based on the performance and cost constraints. Then we exhaustively apply every algorithm-parameter combination to each platform and obtain the corresponding performances. In the run time, we calculate the feature similarity on the manifold that is shared between training and test data. The more similar they are, the higher the possibility that they share the same algorithm-parameter combination.  We obtain the ``best'' algorithm-parameter combination for the test data by selecting the most similar training data's selection. We show the efficacy of the proposed method on the application of pedestrian detection and tracking. Our promising experimental results have demonstrated that adaptively selecting the algorithm-parameter combinations for each scenario is able to obtain the best or is close to the best performance under a certain constraint.

{
\small
 \bibliographystyle{IEEEtran}
 \bibliography{egbib}
}
%\end{thebibliography}

%\begin{biography}[{\includegraphics[width=1in,height=1.25in,clip,keepaspectratio]{mshell}}]{Michael Shell}

%\end{IEEEbiography}
%% insert where needed to balance the two columns on the last page with
%%\newpage


%\vfill


\end{document}